\documentclass[letterpaper]{article} 
\usepackage{aaai2026}  
\usepackage{times}  
\usepackage{helvet}  
\usepackage{courier}  
\usepackage[hyphens]{url}  
\usepackage{graphicx} 
\usepackage{natbib}  
\usepackage{caption} 
\usepackage{algorithm}
\usepackage{algorithmic}
\usepackage{multirow}
\usepackage{subcaption}
\usepackage{amsmath, amsthm, amssymb, amsfonts}
\usepackage{booktabs} 
\usepackage{newfloat}
\usepackage{listings}
\DeclareCaptionStyle{ruled}{labelfont=normalfont,labelsep=colon,strut=off} 
\floatstyle{ruled}
\newfloat{listing}{tb}{lst}{}
\floatname{listing}{Listing}
\title{Uncovering and Mitigating Destructive Multi-Embedding Attacks in Deepfake Proactive Forensics}
\author{
    Lixin Jia\textsuperscript{\rm 1}\equalcontrib,
    Haiyang Sun\textsuperscript{\rm 1}\equalcontrib,
    Zhiqing Guo\textsuperscript{\rm 1,\rm 2}\thanks{Corresponding author: Zhiqing Guo, guozhiqing@xju.edu.cn},
    Yunfeng Diao\textsuperscript{\rm 3},
    Dan Ma\textsuperscript{\rm 1},
    Gaobo Yang\textsuperscript{\rm 4}
}
\affiliations{
    \textsuperscript{\rm 1}School of Computer Science and Technology, Xinjiang University\\
    \textsuperscript{\rm 2}Xinjiang Multimodal Intelligent Processing and Information Security Engineering Technology Research Center\\
    \textsuperscript{\rm 3}School of Computer Science and Information Engineering, Hefei University of Technology\\
    \textsuperscript{\rm 4}College of Computer Science and Electronic Engineering, Hunan University\\
}

\begin{document}

\maketitle

\begin{abstract}
With the rapid evolution of deepfake technologies and the wide dissemination of digital media, personal privacy is facing increasingly serious security threats. Deepfake proactive forensics, which involves embedding imperceptible watermarks to enable reliable source tracking, serves as a crucial defense against these threats. Although existing methods show strong forensic ability, they rely on an idealized assumption of single watermark embedding, which proves impractical in real-world scenarios. In this paper, we formally define and demonstrate the existence of Multi-Embedding Attacks (MEA) for the first time. When a previously protected image undergoes additional rounds of watermark embedding, the original forensic watermark can be destroyed or removed, rendering the entire proactive forensic mechanism ineffective. To address this vulnerability, we propose a general training paradigm named Adversarial Interference Simulation (AIS). Rather than modifying the network architecture, AIS explicitly simulates MEA scenarios during fine-tuning and introduces a resilience-driven loss function to enforce the learning of sparse and stable watermark representations. Our method enables the model to maintain the ability to extract the original watermark correctly even after a second embedding. Extensive experiments demonstrate that our plug-and-play AIS training paradigm significantly enhances the robustness of various existing methods against MEA.
\end{abstract}

\begin{links}
    \link{Code}{https://github.com/vpsg-research/MEA}
\end{links}

\section{Introduction}

\begin{figure}[t]
\centering
\includegraphics[width=\columnwidth]{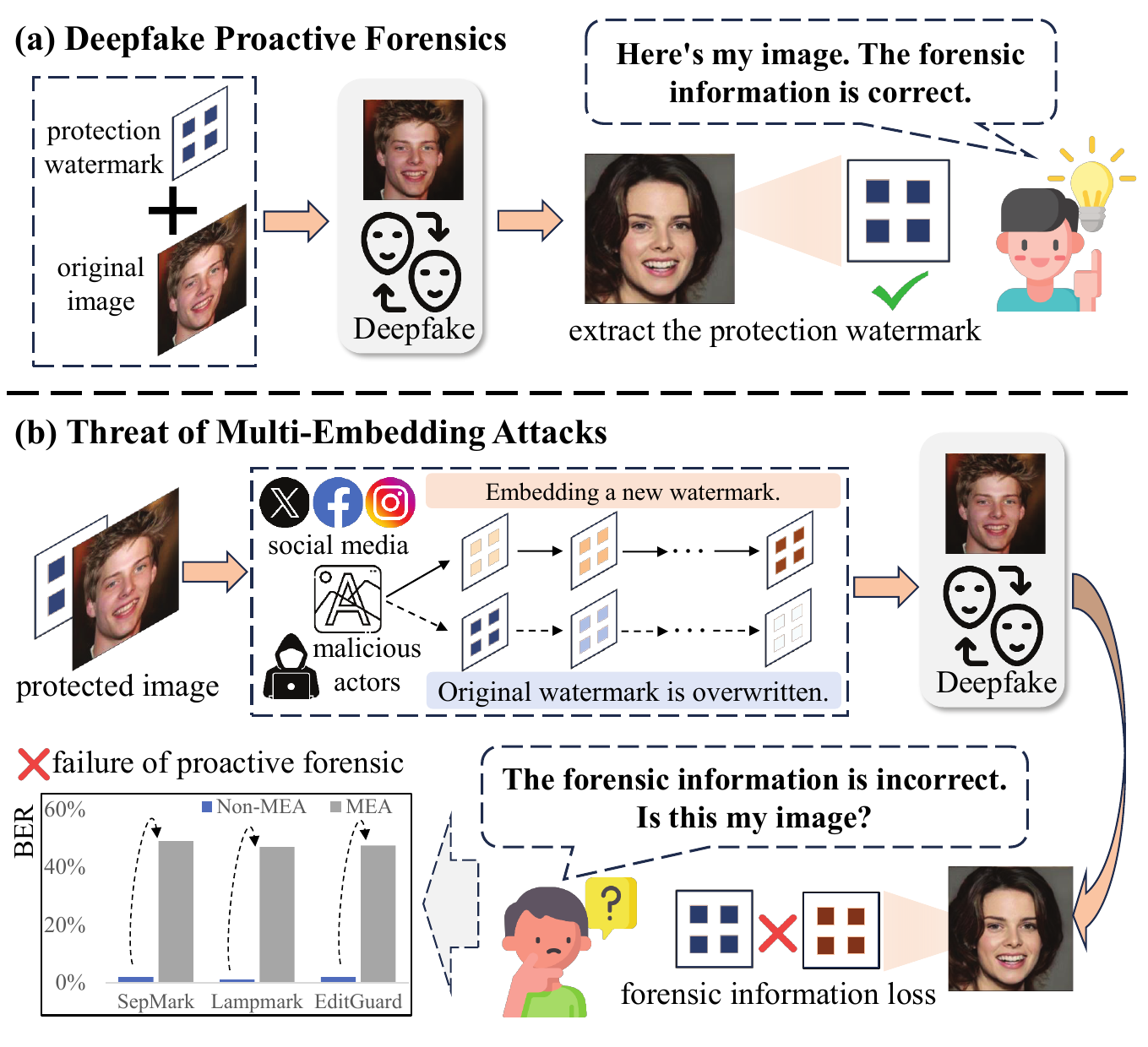}
\caption{Illustration of the ideal proactive forensics pipeline and the threat posed by Multi-Embedding Attacks (MEA). (a) In the standard pipeline, the embedded forensic watermark is expected to withstand manipulations such as deepfake generation. (b) However, additional third-party embeddings (e.g., social media platforms or malicious actors) can overwrite and degrade the original watermark, leading to a complete failure of the proactive forensics mechanism.}
\label{fig1}
\end{figure}

The rise of deepfake technology, particularly with the emergence of face manipulation models such as StarGAN~\cite{choi2020stargan} and LDM~\cite{ldm}, has significantly lowered the barrier to content forgery, leading to severe societal risks. These technologies not only seriously erode the public's trust in digital content, but also manipulate public opinion, infringe on personal privacy and encourage identity fraud. To address this growing challenge, digital forensics is shifting its focus from passive post-hoc detection~\cite{Sun2021ltw, Exploit, Exposing} to proactive forensics~\cite{wu2023sepmark, wang2024lampmark, sun2025diffmark}, aiming to prevent malicious operations at their source. Currently, a leading strategy in deepfake proactive forensics involves embedding invisible digital watermarks~\cite{chang2022blind} into original images, serving as a ``digital fingerprint". When these images are disseminated or even subjected to deepfake manipulation, these embedded watermarks are designed to retain their integrity. This ultimately enables subsequent traceability and copyright protection. This concept is illustrated in Figure~\ref{fig1}(a).

Compared to traditional watermarking techniques~\cite{sy2020traditionwatermarkcnn, wang2021mappingwatermarkcnn}, deep watermarking methods based on encoder–decoder architectures have attracted increasing attention in proactive forensics due to their superior flexibility and embedding performance\cite{jia2021mbrs, neekhara2022faceswatermarks}. Existing approaches have made significant progress in ensuring imperceptibility and robustness of watermarks against distortions~\cite{wang2024robustmark}. However, existing approaches mainly focus on improving the robustness of watermarking schemes against various types of deepfake manipulations and common post-processing (e.g., adding noise), while ignoring the threat of multiple watermark embeddings. Just as facial images are vulnerable to deepfake forgery attacks, they are equally exposed to the risk of multiple watermark embedding operations during their widespread dissemination. Additional watermark embeddings can arise from automated processing by social media platforms. More significantly, malicious actors may deliberately introduce secondary watermarking to obscure the original forensic evidence. This phenomenon leaves a research blank to be solved urgently. That is, how can the forensic information of the original watermark be reliably extracted under the subsequent watermark embedding attack?

In this work, we formally introduce and define the concept of Multi-Embedding Attacks (MEA), which refers to intentionally inserting additional watermarks into already protected images, thereby disrupting the original forensic watermark. This attack scenario is illustrated in Figure~\ref{fig1}(b). To clarify the inherent vulnerability of proactive forensics framework to MEA, we first analyze the existing methods theoretically. Subsequently, through empirical validation of the representative deepfake active forensics methods, we demonstrate that these approaches are highly vulnerable to MEA in practice. The initially embedded forensic watermark suffers severe degradation after undergoing a second round of embedding, ultimately losing its original forensic functionality. This critical limitation poses a fundamental challenge to the security, robustness, and practical deployability of existing proactive forensic techniques.

To this end, we propose Adversarial Interference Simulation (AIS), a general and effective defense paradigm designed to against MEA. Rather than relying on complex architectural designs, AIS adjusts the optimization objective during the fine-tuning stage of the training process. At this stage, the training goal is explicitly defined as resisting interference from subsequent watermark embeddings, thereby preserving the original forensic information. Specifically, we explicitly apply simulated MEA to watermarked images and introduce a resilience loss to guide the encoder to learn more sparse watermark representation. Sparse watermarking has been widely validated as an effective strategy for enhancing robustness against attacks~\cite{Boes2022Securesparse, Deeba2020sparsewatermark}. Building on this foundation, we leverage sparse and stable watermark representations to resist MEA, ensuring the integrity of forensic information under adversarial interference. Crucially, as a model-agnostic training paradigm, AIS can be implemented as a plug-and-play solution to support a wide range of existing and future proactive forensic systems against MEA.

In summary, our main contributions are as follows:
\begin{itemize}
    \item We are the first to define and validate Multi-Embedding Attacks (MEA) in proactive face forensics, a critical and widespread vulnerability that has been overlooked in previous studies. We also reveal that MEA makes existing mainstream proactive forensics methods completely ineffective.
    \item We propose an Adversarial Interference Simulation (AIS), which is a model-agnostic training paradigm designed to resist MEA. AIS simulates adversarial interference during training, using a resilience loss function to promote the learning of robust watermark representations.
    \item Our experiments show that plug-and-play AIS significantly improves the resistance of various state-of-the-art methods to MEA. We also suggest that MEA robustness should be a key benchmark for future research to better meet real-world security demands.
\end{itemize}

\section{Related Works}

Existing passive deepfake detectors often struggle to generalize to unseen manipulation techniques\cite{liu2024faceforgerydetection, xu2024exposingdeepfake, guo2023constructing, LDFnet}, prompting a growing interest in proactive forensic methods~\cite{yu2021fingerprinting, yu2022responsible, he2025kad}. Among these, invisible watermarking based on deep learning has become the dominant paradigm, typically employing an encoder–decoder architecture~\cite{yang2021faceguard}. The robustness of such methods is commonly evaluated against deepfake attacks generated by techniques such as SimSwap~\cite{chen2020simswap}, GANimation~\cite{gao2021information}, and StarGAN~\cite{choi2020stargan}. A variety of sophisticated techniques have been developed to address these known threats, and significant advances have been made. For example, LampMark~\cite{wang2024lampmark} introduces content-aware watermarking guided by facial landmarks to improve resistance. AdvMark~\cite{Wu2024Advmark} adopts adversarial watermarking strategies to improve both passive detection and active traceability under challenging conditions. Furthermore, recent work~\cite{wang2024robustmark} introduces an unpredictable encryption mechanism to strengthen the security and confidentiality of watermark information. Although existing methods perform well in resisting content-level manipulations, they exhibit inherent limitations in both architectural design and optimization objectives. These deficiencies undermine their robustness against structured signal interference introduced by additional watermark embeddings.

In this work, we identify this vulnerability as a common issue in deepfake proactive forensics. Existing methods typically do not train the internal representation associated with the original watermark to resist subsequent targeted overwriting by similar signals. Given the destructive nature of Multi-Embedding Attacks (MEA), there is an urgent need for a generalizable defense strategy. To address this, we simulate MEA during training to guide the model toward learning a sparse and stable watermark representation, thereby enhancing its robustness against such attacks.

\section{Analysis of Multi-Embedding Attacks}
\subsection{Problem Formulation}

In contrast to passive deepfake detection methods, proactive forensics embeds protective watermarks into images in advance to support future identity verification and source attribution. Most current deepfake proactive forensic methods are based on a general encoder-decoder framework. These methods are optimized end-to-end to balance the imperceptibility and robustness of the watermark. A typical proactive forensics pipeline generally comprises four core modules:
\begin{itemize}
    \item \textbf{Encoder ($En$):} Embedding a watermark message $w$ into the original image $X$ to generate a watermarked image $X_w = En(X, w)$.
    \item \textbf{Decoder ($De$):} Recovering the embedded watermark from a potentially distorted watermarked image.
    \item \textbf{Adjustable Noise Layer ($\mathcal{N}$):} Modeling realistic post-processing and adversarial manipulations, such as compression, noise, and deepfake transformations.
    \item \textbf{Discriminator ($\mathcal{G}$):} Guiding the encoder to maintain visual imperceptibility between $X$ and $X_w$.
\end{itemize}

Given an original image $X\in\mathbb{R}^{H \times W \times 3}$ and a protective watermark $w \in \{-\alpha,\alpha\}^L$, the training process is designed to jointly optimize two primary objectives: robustness and imperceptibility. Robustness ensures that the embedded watermark can withstand various distortions, allowing the decoder to accurately recover the original message. Imperceptibility requires that the watermarked image be visually indistinguishable from its original counterpart. The corresponding optimization objective is formally defined as:
\begin{equation}
\label{eq:standard_objective}
\begin{split}
\operatorname*{min}_{En,De} & \mathcal{H}(De(\mathcal{N}(En(X,w))),w)\\
&+\lambda\mathcal{G}(X,En(X,w)).
\end{split}
\end{equation}

The optimization objective minimizes a weighted sum of two terms: a robustness term $\mathcal{H}(De(\cdot), w)$, which measures the watermark recovery error, and an imperceptibility term $\lambda\mathcal{G}(X, \cdot)$, which penalizes perceptual distortions. The core philosophy of this objective is to train the encoder to manipulate pixels in such a way that the resulting watermark perturbations are indistinguishable from natural image noise or textures. However, this standard objective introduces a critical vulnerability. It is built on an implicit and often overlooked assumption that the input image $X$ is a clean, unmodified carrier. The entire training process operates under an idealized scenario where a watermark is embedded only once into an original image. This gap between the training assumption and real-world usage scenarios introduces a critical security threat, which we define as the Multi-Embedding Attacks (MEA). MEA appears when an opponent embeds a second independent watermark in the image that already contains a forensic watermark. This type of attack directly exploits the lack of robustness in existing proactive forensic frameworks against multiple embedding operations, revealing a previously unaddressed vulnerability in their design.

\begin{figure}[t]
    \centering
    \begin{subfigure}[t]{0.48\columnwidth}
        \centering
        \includegraphics[width=\linewidth]{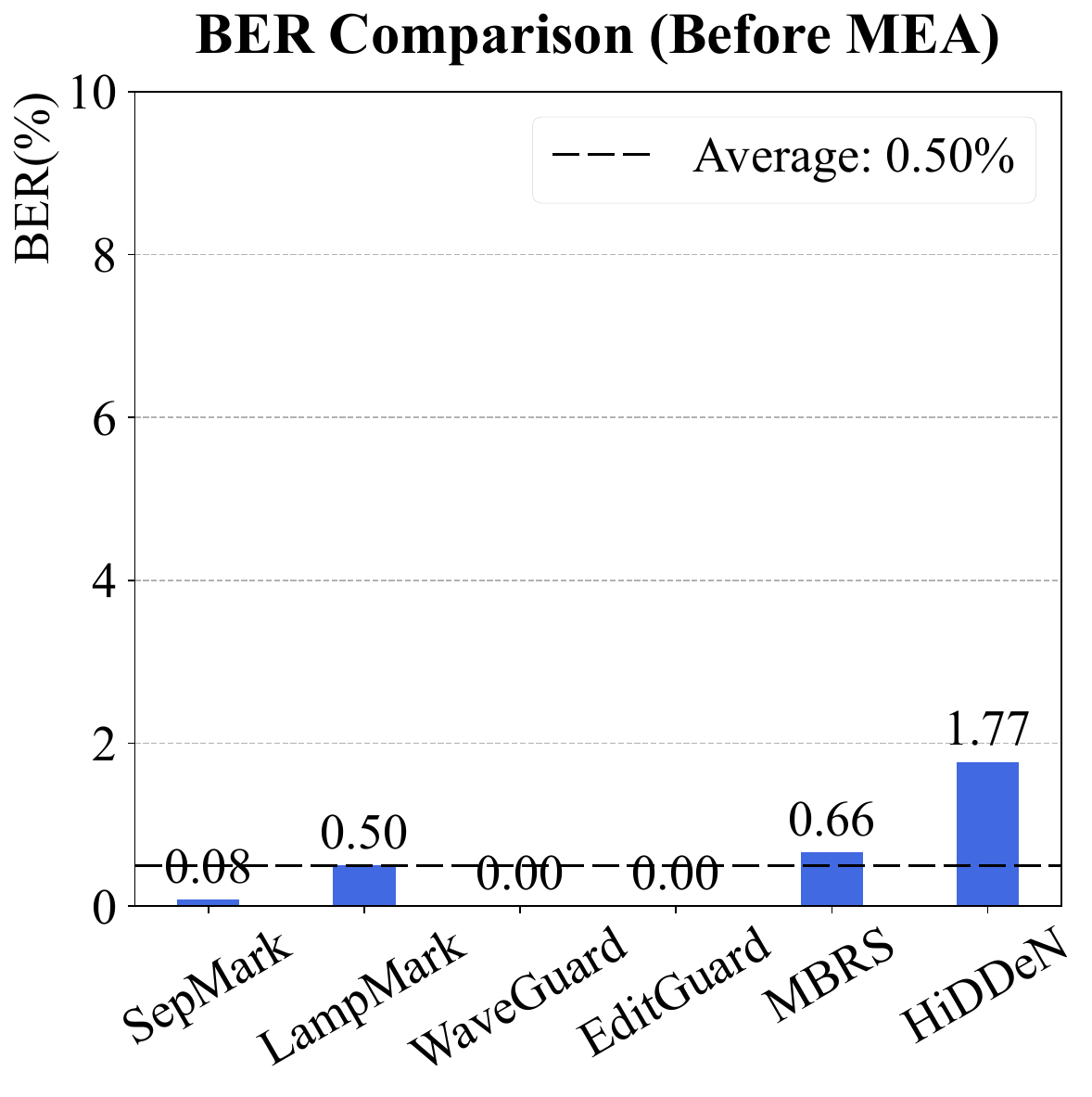}
        \label{fig:ber_before}
    \end{subfigure}
    \hfill
    \begin{subfigure}[t]{0.48\columnwidth}
        \centering
        \includegraphics[width=\linewidth]{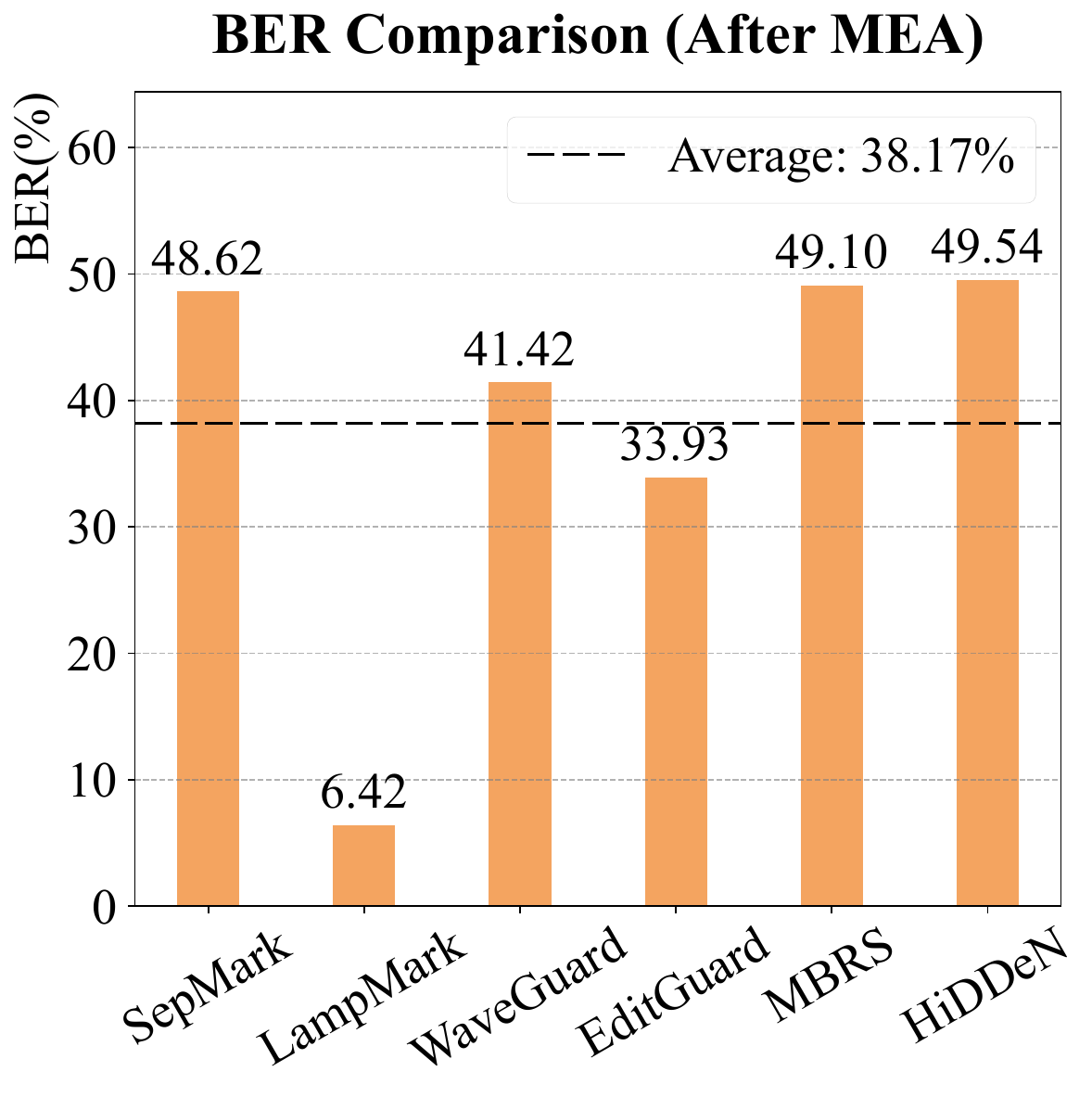}
        \label{fig:ber_after}
    \end{subfigure}
    \caption{BER of primary watermarks before and after MEA on several state-of-the-art methods.}
    \label{fig2}
\end{figure}

\subsection{Theoretical Analysis of MEA}
\label{subsec:assumption_break}

The vulnerability of existing frameworks to MEA is a direct consequence of the optimization objective defined in Eq.~\ref{eq:standard_objective}. Consider the MEA process: an adversary, or a third-party service, takes a protected image $X_{w_1} = En(X, w_1)$ and embeds a new watermark $w_2$. This second embedding process, following the same standard paradigm, solves an analogous optimization problem:
\begin{equation}
\label{eq:mea_objective}
\begin{split}
\operatorname*{min}_{En,De}
&\mathcal{H}(De(\mathcal{N}(En(X_{w_1},w_2))),w_2) \\ 
&+ \lambda\mathcal{G}(X_{w_1},En(X_{w_1},w_2)).
\end{split}
\end{equation}

\begin{figure*}[t]
\centering
\includegraphics[width=\textwidth]{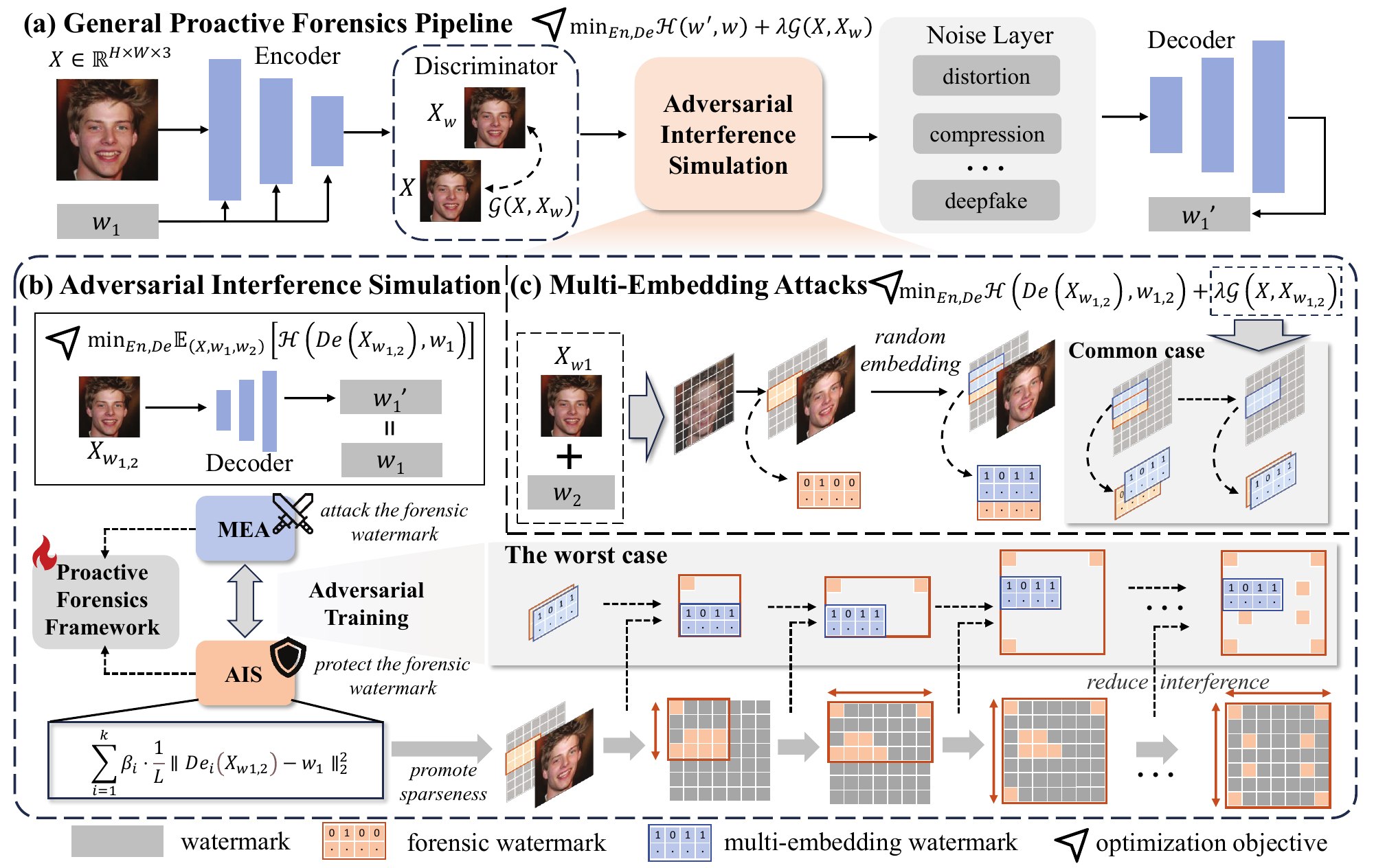}
\caption{Overview of the proposed framework. Figure (a) illustrates the general proactive forensics pipeline. Figure (b) depicts our proposed Adversarial Interference Simulation (AIS), a model-agnostic training paradigm applicable to various proactive forensic methods. Figure (c) shows the Multi-Embedding Attacks (MEA), where the invisibility constraint of watermark embedding typically causes new watermarks to overwrite the original forensic information.}
\label{fig3}
\end{figure*}

The critical component is the imperceptibility term $\mathcal{G}(X_{w_1},En(X_{w_1},w_2))$. To minimize this term while embedding the new message $w_2$, the encoder is optimized to make the smallest possible changes to its input $X_{w_1}$. As a result, when the second embedding occurs, the new encoder also attempts to embed $w_2$ while minimizing visual distortion, as illustrated in Figure~\ref{fig3}(c). During this process, subtle pixel-level perturbations of the original watermark $w_1$ embedded in $X_{w_1}$ are often misinterpreted as redundant textures or noise. This misinterpretation is inevitable because these perturbations are intentionally designed to be indistinguishable from the intrinsic features of the image to remain imperceptible. Consequently, the original forensic watermark is not preserved but is instead targeted and overwritten, guaranteeing the attack's success and rendering the initial protection useless.

From a forensic standpoint, the success of this attack is the destruction of the original evidence. We quantify this by the loss of forensic integrity $\mathcal{L}_f$, which measures the recovery error of the original watermark $w_1$ from the doubly watermarked image $X_{w_{1,2}}$:
\begin{equation}
\label{eq:forensic_loss}
\mathcal{L}_f = \mathcal{H}(De(X_{w_{1,2}}), w_{1}).
\end{equation}
Our theoretical analysis predicts that for any method adhering to the standard paradigm, a successful MEA will result in a high value for $\mathcal{L}_f$, rendering the original watermark irrecoverable. In practice, this high loss corresponds to a high bit error rate during decoding.

\subsection{Empirical Validation of MEA}
To validate our theoretical analysis, we provide empirical evidence supporting our claim that MEA constitutes a general vulnerability inherent in existing proactive forensic models, including SepMark~\cite{wu2023sepmark}, LampMark~\cite{wang2024lampmark}, WaveGuard~\cite{he2025waveguard}, EditGuard~\cite{zhang2024editguard}. In addition, due to the limited availability of open-source proactive forensic methods, we include two representative end-to-end robust watermarking methods, MBRS~\cite{jia2021mbrs} and HiDDeN~\cite{zhu2018hidden}, in our evaluation.

The experimental procedure consists of three core steps. (1) For each image $X$ in the test set, a unique original watermark $w_1$ embedded using a pretrained encoder $En_k$, resulting in the protected image $X_{w_1}$. (2) A MEA is simulated by re-encoding $X_{w_1}$ with the same encoder $E_k$, embedding a new unrelated random watermark $w_2$ to produce the rewritten image $X_{w_{1,2}}$. (3) Finally, the decoder $De_k$ attempts to recover the original watermark $w_1$ from $X_{w_{1,2}}$, and the bit error rate (BER) between $w_1$ and the recovered $w_1^{\prime}$ is calculated. The BER is defined as:
\begin{equation}
\label{eq:ber}
\mathrm{BER}(w_1,w_1^\prime)=\frac{\sum_{i=0}^{L-1}\left|\mathcal{B}(w_1^i)-\mathcal{B}(w_1^{\prime i})\right|}{L}\times100\%,
\end{equation}
where $L$ is the message length and $\mathcal{B}(\cdot)$ denotes a binarization function for mapping positive values to 1 and non-positive values to 0.

The results in Figure~\ref{fig2} clearly support our analysis. After MEA, the average BER jumps from 0.50\% to 38.17\%, indicating substantial degradation across all evaluated methods. Among them, LampMark remains notably robust due to its landmark-based embedding strategy, which inherently resists spatial overwriting. In contrast, methods such as SepMark, MBRS, and HiDDeN experience BER approaching 50\%. In binary data communication, a BER approaching 50\% is statistically equivalent to random guessing, indicating a complete loss of the original forensic information. This outcome provides conclusive empirical evidence that the vulnerability to MEA is not an isolated implementation flaw but a systemic weakness stemming from the foundational design of the current proactive forensics paradigm. These findings emphasize the urgent need to establish a new defense paradigm against MEA.

\section{Methodology}
Motivated by the systemic vulnerability to rewriting attacks, we propose Adversarial Interference Simulation (AIS) to ensure the robustness of forensic information by simulating MEA interference during training. To ensure a fair and direct comparison with existing works, our method is built upon the same encoder-decoder architecture and utilizes the foundational configurations outlined in Motivation section.

\subsection{Adversarial Interference Simulation}

\subsubsection{Learning Sparse Representations.} 
Our analysis reveals that the vulnerability of proactive forensic models to MEA arises from the dense watermark representation produced by standard end-to-end training. These representations lack a clear distinction, which makes the embedded forensic watermark fragile and easily covered by subsequent embedding operations. To address this, we introduce Adversarial Interference Simulation (AIS), a training paradigm that encourages the formation of sparse and interference-resilient watermark features. In such a representation, the forensic information is concentrated within a compact, stable, and decoupled subset of feature dimensions, showing strong resistance to subsequent embedding interference.

Therefore, the central idea of AIS lies in its redefinition of the optimization process. Our AIS training paradigm significantly improves the model's ability to resist MEA by simulating adversarial scenarios during training, rather than merely learning to recover watermarks after standard distortions, as illustrated in Figure~\ref{fig3}(b). This objective fundamentally alters the optimization landscape, guiding it to actively mitigate interference from subsequent watermark embeddings during training. Let the initial protected image be $X_{w_1}=En(X, w_1)$. This image then passes through a simulated distortion channel, resulting in $\tilde{X}_{w_1}=\mathcal{N}(X_{w_1})$. Subsequently, we simulate the attack by embedding a second random watermark $w_2$, producing the doubly watermarked image $X_{w_{1,2}} = En(\tilde{X}_{w_1}, w_2)$. The core objective of AIS is to train the encoder $En$ and decoder $De$ to minimize the expected recovery error of the original watermark $w_1$ from the attacked image $X_{w_{1,2}}$. This is formally defined as:
\begin{equation}
\label{eq:AIS_objective}
\min_{En,De}\mathbb{E}_{(X,w_1,w_2)}\left[\mathcal{H}\left(De(X_{w_{1,2}}),w_1\right)\right].
\end{equation}
AIS encourages the model to embed w1 sparsely, concentrating critical information in stable, less interfered dimensions. This method effectively reduces the risk of information being overwritten in the worst case, significantly improving the robustness of the watermark against destructive interference caused by secondary embedding operations.

\subsubsection{Loss Formulation.} To implement the AIS training paradigm, we integrate our objective into the existing end-to-end training pipeline of a proactive forensics model. The total loss function $\mathcal{L}_{Total}$ is a carefully weighted composite of two primary components: a standard task loss $\mathcal{L}_{task}$ and our proposed resilience loss $\mathcal{L}_{AIS}$. 

The standard task Loss ensures the model's efficacy for a single watermark embedding. While specific components may vary slightly across different baseline architectures (e.g., SepMark), it consistently encompasses three fundamental objectives: (1) an encoder loss $\mathcal{L}_{I}$ for imperceptibility; (2) a multi-headed decoder loss $\mathcal{L}_{D}$ for basic robustness; (3) an adversarial loss $\mathcal{L}_{A}$ to further enhance stealthiness. The general form of the task loss is therefore defined as:
\begin{equation}
\label{eq:Ltask}
\mathcal{L}_{task} = \lambda_{1}\mathcal{L}_{I} + \lambda_{2}\mathcal{L}_{D} + \lambda_{3}\mathcal{L}_{A}.
\end{equation}

The core of our innovation lies in the resilience loss $\mathcal{L}_{AIS}$. This loss directly operationalizes our high-level objective from Eq. \ref{eq:AIS_objective} and is designed to be agnostic to the specific decoder architecture. Let $\{De_i\}_{i=1}^k$ denote the set of all decoder components within the baseline framework responsible for watermark extraction (where $k\geq1$). Our robustness objective requires that each decoder in this set accurately recover the original watermark $w_1$ from the image $X_{w_{1,2}}$, which has been subjected to the MEA. For example, in a typical single-decoder model, we have $k = 1$. In contrast, architectures like SepMark adopt a multi-head design that includes multiple primary decoders for $w_1$ recovery, as well as an auxiliary decoder $De_0$ trained to output a zero vector. The resulting $\mathcal{L}_{AIS}$ is thus formulated as:
\begin{equation}
\begin{split}
\label{eq:LAIS_expanded}
\mathcal{L}_{AIS} = &\sum_{i=1}^k \beta_i \cdot \frac{1}{L} \| De_i(X_{w{1,2}}) - w_1 \|_2^2 \\
&+ \beta_0 \cdot \frac{1}{L} \| De_0(X_{w{1,2}}) - \textbf{0} \|_2^2.
\end{split}
\end{equation}
We use MSE loss, normalized by message length $L$, to measure the discrepancy between decoded and original messages. The weighting coefficients $\beta_k$ and $\beta_0$ regulate the contribution of each decoder head to the overall loss. The coefficients vary with the forensic framework. When the auxiliary decoder $De_0$ is absent, $\beta_0$ is assigned a value of 0.

Finally, the total optimization objective for our AIS framework integrates the resilience term with the standard task loss:
\begin{equation}
\label{Ltotal}
\mathcal{L}_{Total} = \mathcal{L}_{task} + \lambda_A\mathcal{L}_{AIS}.
\end{equation}
By minimizing the joint objective, the $\mathcal{L}_{AIS}$ term acts as a strong sparsity-inducing regularizer. It guides the framework to learn an embedding strategy that remains effective in single-use scenarios while also being inherently robust against the critical threat of MEA, by promoting a sparse and resilient representational structure.

\begin{table*}[!ht]
\centering
\small 
\setlength{\tabcolsep}{1mm}

\begin{tabular*}{\textwidth}{c c |@{\extracolsep{\fill}} *{6}{c}}
\toprule
\textbf{Metric} & \textbf{Emb.} & \textbf{SepMark} & \textbf{LampMark} & \textbf{WaveGuard} & \textbf{EditGuard} & \textbf{MBRS} & \textbf{HiDDeN} \\
\midrule
\multirow{5}{*}{\begin{tabular}{c}\textbf{BER (\%)}\\ $\downarrow$\end{tabular}} 
& 1 & 0.0775 / \textbf{0.0227} & 0.5034 / \textbf{0.1135} & 0.0000 / 0.0000 & 0.0000 / 0.0068 & 0.6629 / \textbf{0.4869} & 1.8827 / \textbf{1.7747} \\
& 2 & 48.6223 / \textbf{0.0465} & 6.4235 / \textbf{0.1196} & 41.4232 / \textbf{0.0211} & 33.9345 / \textbf{2.4283} & 49.0960 / \textbf{1.5239} & 49.5370 / \textbf{2.0679} \\
& 3 & 50.0094 / \textbf{0.0459} & 12.0529 / \textbf{0.1383} & 25.0035 / \textbf{0.2376} & 23.9296 / \textbf{2.3058} & 47.0838 / \textbf{1.9526} & 49.8148 / \textbf{1.6667} \\
& 4 & 50.0022 / \textbf{0.0681} & 16.0535 / \textbf{0.1706} & 41.1283 / \textbf{1.8165} & 35.3829 / \textbf{2.3914} & 48.2058 / \textbf{2.2891} & 49.4290 / \textbf{1.3426} \\
& 5 & 50.1732 / \textbf{0.1123} & 18.6007 / \textbf{0.2213} & 34.7835 / \textbf{5.2587} & 34.7820 / \textbf{2.2408} & 48.1710 / \textbf{2.5076} & 50.1389 / \textbf{1.9599} \\[-1pt]
\midrule
\multirow{5}{*}{\begin{tabular}{c}\textbf{PSNR}\\ $\uparrow$\end{tabular}} 
& 1 & 42.1015 / \textbf{42.4959} & 27.0208 / \textbf{28.6729} & 37.0220 / 33.4610 & 42.1853 / \textbf{57.2217} & 38.4673 / 37.8905 & 25.2963 / 22.0898 \\
& 2 & 35.2802 / \textbf{36.2940} & 23.9187 / \textbf{26.5957} & 35.1625 / 30.9346 & 38.2056 / \textbf{50.5427} & 23.4829 / 20.0247 & 22.9553 / 17.3593 \\
& 3 & 33.3797 / \textbf{35.0832} & 21.3854 / \textbf{24.1853} & 33.9053 / 29.3663 & 35.7159 / \textbf{44.3066} & 23.1079 / 18.6029 & 21.4339 / 16.1444 \\
& 4 & 31.7767 / \textbf{33.8606} & 19.4583 / \textbf{22.0994} & 33.1148 / 28.1904 & 33.9585 / \textbf{40.6596} & 22.7785 / 17.4066 & 20.8483 / 15.0353 \\
& 5 & 30.5249 / \textbf{32.5170} & 17.9343 / \textbf{20.3220} & 32.5401 / 27.3020 & 32.5557 / \textbf{38.2171} & 22.5302 / 16.4738 & 18.9704 / 14.4961 \\[-1pt]
\midrule
\multirow{5}{*}{\begin{tabular}{c}\textbf{SSIM}\\ $\uparrow$\end{tabular}} 
& 1 & 0.9730 / 0.9641 & 0.8834 / \textbf{0.8984} & 0.8970 / 0.8325 & 0.9910 / \textbf{0.9990} & 0.9340 / 0.9267 & 0.8689 / \textbf{0.8718} \\
& 2 & 0.8917 / \textbf{0.9010} & 0.7644 / \textbf{0.8692} & 0.8791 / 0.8011 & 0.9820 / \textbf{0.9964} & 0.7044 / 0.4501 & 0.7958 / 0.7925 \\
& 3 & 0.8522 / \textbf{0.8937} & 0.6750 / \textbf{0.8368} & 0.8600 / 0.7778 & 0.9717 / \textbf{0.9902} & 0.6808 / 0.4053 & 0.7450 / \textbf{0.7966} \\
& 4 & 0.8067 / \textbf{0.8823} & 0.6021 / \textbf{0.8042} & 0.8486 / 0.7621 & 0.9608 / \textbf{0.9808} & 0.6572 / 0.3640 & 0.7324 / \textbf{0.7510} \\
& 5 & 0.7639 / \textbf{0.8665} & 0.5416 / \textbf{0.7725} & 0.8399 / 0.7511 & 0.9488 / \textbf{0.9684} & 0.6308 / 0.3386 & 0.6754 / \textbf{0.7329} \\[-1pt]
\bottomrule
\end{tabular*}
\caption{Performance comparison under intra-model MEA. Each baseline model is evaluated alongside its AIS-enhanced version (+AIS), with Emb. denoting the number of embedding operations. Results are reported in the format (Baseline / +AIS), and bold values indicate superior performance by the AIS-enhanced model.}
\label{tab:performance_comparison_double}
\end{table*}

\begin{table*}[t]
\centering
\small

\setlength{\tabcolsep}{1mm}
\begin{tabular*}{\textwidth}{c @{\extracolsep{\fill}} c *{5}{c}}
\toprule
\multicolumn{2}{c}{\textbf{Model}} & \multicolumn{5}{c}{\textbf{Additional Embeddings}} \\
\cmidrule(lr){1-2} \cmidrule(lr){3-7}
\textbf{Defender} & \textbf{Attacker} & \textbf{n=1} & \textbf{n=2} & \textbf{n=3} & \textbf{n=4} & \textbf{n=5} \\
\midrule
\multirow{3}{*}{\textbf{SepMark}} 
& LampMark  & 5.59\%\,/\,0.26\% & 9.74\%\,/\,0.24\% & 16.69\%\,/\,0.48\% & 21.59\%\,/\,0.75\% & 25.74\%\,/\,1.31\% \\
& EditGuard & 1.22\%\,/\,0.15\% & 1.23\%\,/\,0.17\% & 1.38\%\,/\,0.20\% & 1.62\%\,/\,0.20\% & 1.58\%\,/\,0.15\% \\
& HiDDeN    & 2.48\%\,/\,0.27\% & 2.67\%\,/\,0.41\% & 3.17\%\,/\,0.64\% & 3.75\%\,/\,0.66\% & 4.15\%\,/\,0.75\% \\[-2pt]
\midrule
\multirow{3}{*}{\textbf{LampMark}} 
& SepMark   & 3.85\%\,/\,2.87\% & 6.87\%\,/\,5.70\% & 8.21\%\,/\,6.60\% & 8.55\%\,/\,7.79\% & 11.05\%\,/\,10.04\% \\
& EditGuard & 0.51\%\,/\,0.11\% & 0.55\%\,/\,0.11\% & 0.57\%\,/\,0.12\% & 0.62\%\,/\,0.12\% & 0.67\%\,/\,0.13\% \\
& HiDDeN    & 0.68\%\,/\,0.19\% & 0.83\%\,/\,0.17\% & 1.03\%\,/\,0.16\% & 1.37\%\,/\,0.17\% & 1.45\%\,/\,0.18\% \\[-2pt]
\midrule
\multirow{3}{*}{\textbf{EditGuard}} 
& SepMark   & 50.51\%\,/\,23.17\% & 47.15\%\,/\,20.25\% & 55.92\%\,/\,22.90\% & 55.62\%\,/\,22.75\% & 46.69\%\,/\,22.87\% \\
& LampMark  & 37.77\%\,/\,1.14\% & 40.78\%\,/\,2.88\% & 39.90\%\,/\,5.62\% & 43.93\%\,/\,8.44\% & 45.01\%\,/\,12.32\% \\
& HiDDeN    & 29.78\%\,/\,0.09\% & 29.09\%\,/\,0.09\% & 30.08\%\,/\,0.06\% & 28.58\%\,/\,0.12\% & 31.67\%\,/\,0.06\% \\[-2pt]
\bottomrule
\end{tabular*}
\caption{Generalization performance against cross-model MEA. Each cell reports the BER in the format (Baseline / +AIS).}
\label{tab:cross_model_robustness}
\end{table*}

\begin{figure*}[t]
\centering
\includegraphics[width=0.9\textwidth]{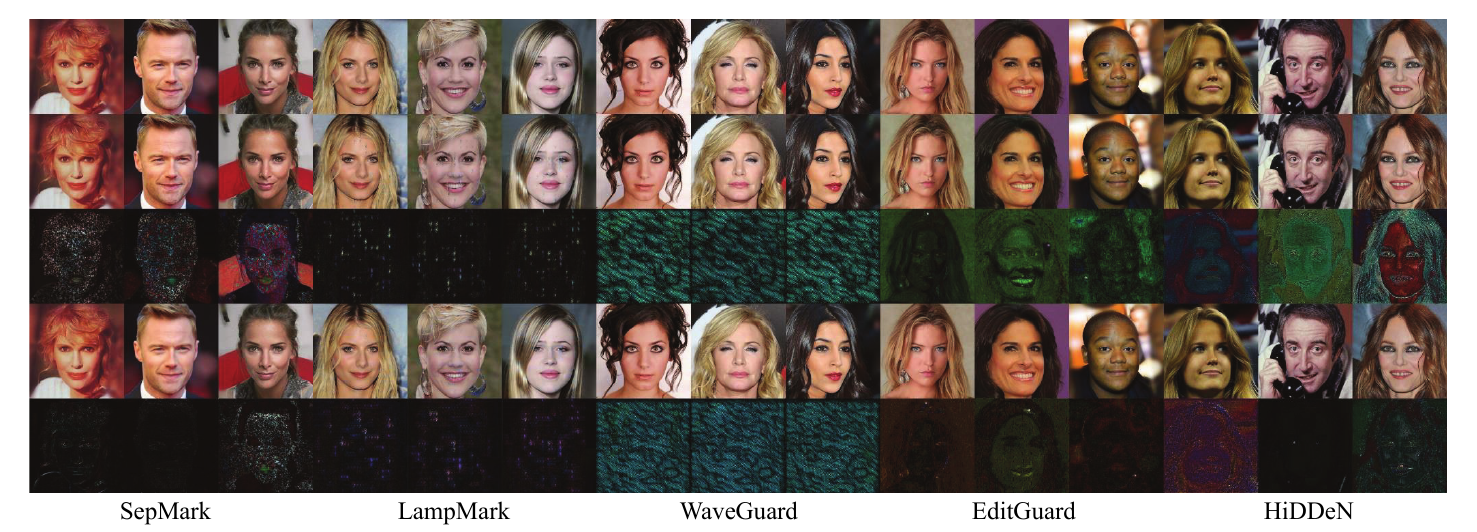}
\caption{Visual comparison of different models before and after enhancement with AIS. From top to bottom are the original image $X$, the watermarked image $X_w$, the residual signal $\mathcal{R}(|X_w - X|)$, the watermarked image after fine-tuning with AIS $X_w^\prime$, and its residual $\mathcal{R}(|X_w^\prime - X|)$. To visualize the minute differences introduced by the watermark, the residual signal is amplified using the normalization function $\mathcal{R}(X) = (X - min(X)) / (max(X) -min(X))$. Image size: 256 × 256.}
\label{fig4}
\end{figure*}

\begin{table*}[t]
\centering
\small

\begin{tabular}{c | *{5}{c}}
\toprule
\textbf{Weight} & \textbf{Emb.=1} & \textbf{Emb.=2} & \textbf{Emb.=3} & \textbf{Emb.=4} & \textbf{Emb.=5} \\
\midrule
\textbf{$ \lambda_A=0$} & 0.50\%\,/\,27.02\,/\,0.87 & 6.42\%\,/\,23.92\,/\,0.76 & 12.06\%\,/\,21.39\,/\,0.67 & 16.05\%\,/\,19.46\,/\,0.60 & 18.60\%\,/\,17.93\,/\,0.54 \\
\textbf{$ \lambda_A=2$} & 0.21\%\,/\,29.26\,/\,0.91 & 0.24\%\,/\,27.11\,/\,0.88 & 0.28\%\,/\,24.66\,/\,0.85 & 0.35\%\,/\,22.57\,/\,0.82 & 0.45\%\,/\,20.80\,/\,0.78 \\
\textbf{$ \lambda_A=4$} & 0.17\%\,/\,29.09\,/\,0.91 & 0.18\%\,/\,27.01\,/\,0.88 & 0.21\%\,/\,24.60\,/\,0.85 & 0.26\%\,/\,22.52\,/\,0.82 & 0.33\%\,/\,20.77\,/\,0.78 \\
\textbf{$ \lambda_A=6$} & 0.14\%\,/\,28.92\,/\,0.90 & 0.15\%\,/\,26.92\,/\,0.87 & 0.17\%\,/\,24.59\,/\,0.84 & 0.21\%\,/\,22.56\,/\,0.81 & 0.27\%\,/\,20.84\,/\,0.78 \\
\textbf{$ \lambda_A=8$} & 0.12\%\,/\,28.77\,/\,0.90 & 0.13\%\,/\,26.79\,/\,0.87 & 0.15\%\,/\,24.47\,/\,0.84 & 0.19\%\,/\,22.45\,/\,0.81 & 0.24\%\,/\,20.72\,/\,0.78 \\
\textbf{$ \lambda_A=10$} & 0.11\%\,/\,28.67\,/\,0.90 & 0.12\%\,/\,26.60\,/\,0.87 & 0.14\%\,/\,24.19\,/\,0.84 & 0.17\%\,/\,22.10\,/\,0.80 & 0.22\%\,/\,20.32/0.77 \\[-2pt]
\bottomrule
\end{tabular}
\caption{Ablation study on the AIS loss weight $\lambda_A$ for the LampMark model (Metric: BER $\downarrow$ / PSNR $\uparrow$ / SSIM $\uparrow$).}
\label{tab:ablation}
\end{table*}

\section{Experiments}

\subsection{Datasets and Implementation Details}

\subsubsection{Dataset and Baselines.} We conducted experiments on the the widely used CelebA-HQ~\cite{karras2018celebhq} dataset (256×256), and further performed cross-dataset evaluation on the LFW~\cite{huang2008lfw} dataset. To comprehensively evaluate the threat posed by MEA and the generalizability of our proposed AIS paradigm, we select a set of representative baseline methods, including SepMark~\cite{wu2023sepmark}, LampMark~\cite{wang2024lampmark}, WaveGuard~\cite{he2025waveguard}, EditGuard~\cite{zhang2024editguard}, MBRS~\cite{jia2021mbrs}, and HiDDeN~\cite{zhu2018hidden}.

\subsubsection{Implementation details.} All experiments were carried out using the PyTorch~\cite{paszke2019pytorch} framework on NVIDIA RTX 3090. The Adam optimizer~\cite{kingma2014adam} is used in all settings. During AIS fine-tuning, we follow the original hyperparameter settings of each baseline to ensure consistency. Detailed hyperparameter settings and training schedules for all models can be found in appendix.

\subsubsection{Evaluation metrics.} We assess performance using a comprehensive set of metrics. The primary indicator of robustness is the bit error rate (BER), which measures the integrity of the recovered forensic information; lower BER values indicate more successful recovery. To evaluate the visual quality of the watermarked images, we use the average peak signal-to-noise ratio (PSNR) and the structural similarity index measure (SSIM) \cite{wang2004ssim}. These metrics quantify signal distortion and structural fidelity, respectively, with higher values reflecting less perceptible degradation caused by watermark embeddings.

\subsection{Robustness to Intra-model MEA}
To simulate scenarios in which adversaries deliberately disrupt forensic watermarks, we consider Intra-model MEA, a form of attack that poses significant threats to forensic methods.
Table~\ref{tab:performance_comparison_double} validates the effectiveness the proposed AIS paradigm. Without AIS, the BER of baseline models rises sharply under repeated embeddings, approaching 50\% and indicating a near-complete failure of forensic functionality. In contrast, models enhanced with AIS consistently maintain low BERs even after five successive attacks, showing strong resilience. This robustness is achieved without sacrificing visual quality, as indicated by consistently high PSNR and SSIM scores. These results indicate that AIS can be seamlessly integrated into existing methods as a plug-and-play module that markedly improves robustness against MEA.
To examine cross-dataset generalization, we also evaluate the LTW dataset, with results presented in Table~\ref{tab:performance_ltw}.

\begin{table}[t]
\centering
\small
\setlength{\tabcolsep}{1mm}
\begin{tabular*}{\linewidth}{@{\extracolsep{\fill}} c c c c c}
\toprule
\textbf{Metric} & \textbf{Emb.} & \textbf{SepMark} & \textbf{EditGuard} & \textbf{MBRS} \\[-2pt]
\midrule
\multirow{2}{*}{\begin{tabular}{c}\textbf{BER (\%)} $\downarrow$\end{tabular}}
& 1 & 0.49 / 3.17 & 0.02 / 8.29 & 0.20 / 0.58 \\
& 2 & 49.27 / 20.16 & 43.41 / 12.93 & 48.26 / 1.05 \\[-2pt]
\midrule
\multirow{2}{*}{\begin{tabular}{c}\textbf{PSNR} $\uparrow$\end{tabular}}
& 1 & 25.50 / 25.98 & 42.57 / 48.12 & 38.06 / 38.01 \\
& 2 & 13.29 / 11.29 & 38.62 / 43.76 & 34.73 / 34.38 \\[-2pt]
\midrule
\multirow{2}{*}{\begin{tabular}{c}\textbf{SSIM} $\uparrow$\end{tabular}}
& 1 & 0.69/ 0.71 & 0.99 / 1.00 & 0.93 / 0.92 \\
& 2 & 0.22 / 0.15 & 0.98 / 0.99 & 0.88 / 0.88 \\[-2pt]
\bottomrule
\end{tabular*}
\caption{Performance comparison on the LFW dataset. Results are reported in the format (Baseline / +AIS).}
\label{tab:performance_ltw}
\end{table}

\subsection{Generalization to Cross-model MEA}
This experiment simulates a scenario in which protected images are additionally embedded with new watermarks by various unknown third-party models. Although these additional embeddings are not explicitly designed to disrupt forensic information, they still cause BER degradation. Table~\ref{tab:cross_model_robustness} reports the results for three models, each showing varying levels of vulnerability to MEA due to differences in their invisible watermarking strategies. EditGuard is particularly fragile without AIS, with BER exceeding 50\% under additional embeddings. While SepMark and LampMark exhibit relatively stronger robustness, their performance still deteriorates as the number of embeddings increases.

\subsection{Ablation Study}

Ablation studies validating the effectiveness of our proposed method are shown in Table~\ref{tab:performance_comparison_double}. The results clearly demonstrate that any forensic model without AIS fails completely under MEA. To assess the contribution of the resilience loss $\mathcal{L}_{AIS}$, we conduct a controlled ablation by varying its weight $\lambda_A$, as shown in Table~\ref{tab:ablation}. The baseline configuration ($\lambda_A=0$), which excludes AIS, exhibits poor performance, with the BER exceeding 18\%. As $\lambda_A$ increases, the robustness improves significantly. Notably, at $\lambda_A=10$, the BER drops below 0.25\% even after five successive attacks. Although this setting results in a slight degradation in visual quality (PSNR/SSIM), the gain in forensic robustness is significant. These results confirm that $\mathcal{L}_{AIS}$ is critical for resisting MEA, and we adopt $\lambda_A = 10$ for all main experiments.

\subsection{Visual Quality Analysis}
AIS maintains excellent visual quality, confirmed by high PSNR and SSIM scores in Table~\ref{tab:performance_comparison_double}. Figure~\ref{fig4} provides a direct qualitative confirmation. We can clearly observe that the watermark images generated by all methods are visually indistinguishable from the original images, whether they are enhanced by AIS or not. This proves that our AIS paradigm significantly improves the security without sacrificing the most basic imperceptibility requirement of the watermark, which is completely consistent with the conclusion of objective indicators. Due to varying watermark embedding strategies, the magnified residuals exhibit distinct visual patterns.

\section{Conclusion}
In this work, we uncover Multi-Embedding Attacks (MEA) as a critical vulnerability in proactive deepfake forensics and propose Adversarial Interference Simulation (AIS) as a general and effective defense strategy. AIS is a model-agnostic training paradigm that significantly improves robustness by encouraging models to learn sparse and interference-resilient representations. Extensive experiments have validated that AIS can be seamlessly integrated as a plug-and-play module to protect various state-of-the-art proactive methods against MEA without sacrificing visual quality. In addition, considering that MEA poses a great threat to proactive forensics methods, we propose robustness against MEA as a new benchmark task for future research. We also advocate for developing more effective defense strategies to mitigate such attacks.

\section{Acknowledgments}
This work was supported in part by the National Natural Science Foundation of China under Grant 62462060, Grant 62302427, and Grant 62472368, in part by the Natural Science Foundation of Xinjiang Uygur Autonomous Region under Grant 2023D01C175.

\bibliography{aaai2026}

\end{document}